\documentclass[11pt,a4paper]{article}
\usepackage[hyperref]{emnlp2020}
\usepackage{times}
\usepackage{latexsym}
\usepackage{multirow}
\usepackage{graphicx}
\usepackage{makecell}
\usepackage{amssymb}
\usepackage{pifont}

\usepackage{microtype}

\aclfinalcopy 
\def\aclpaperid{55} 

\title{CheXbert: Combining Automatic Labelers and Expert Annotations for Accurate Radiology Report Labeling Using BERT
}
\author{Akshay Smit     \textsuperscript{*}\\
  Stanford University \\
  \texttt{\small akshaysm@stanford.edu} \\\And
  Saahil Jain  \textsuperscript{*}\\
  Stanford University \\
  \texttt{\small saahil.jain@stanford.edu} \\\And
  Pranav Rajpurkar \textsuperscript{*}\\
  Stanford University \\
  \texttt{\small pranavsr@cs.stanford.edu} \\ \AND 
  Anuj Pareek \\ 
  Stanford University \\
  \\ \And 
  Andrew Y. Ng \\
  Stanford University \\
  \\ \And 
  Matthew P. Lungren \\
  Stanford University \\
  }
\date{}

\begin{document}
\maketitle

\begingroup\renewcommand\thefootnote{*}
\footnotetext{Equal contribution}
\endgroup

\begin{abstract}
The extraction of labels from radiology text reports enables large-scale training of medical imaging models. Existing approaches to report labeling typically rely either on sophisticated feature engineering based on medical domain knowledge or manual annotations by experts. In this work, we introduce a BERT-based approach to medical image report labeling that exploits both the scale of available rule-based systems and the quality of expert annotations. We demonstrate superior performance of a biomedically pretrained BERT model first trained on annotations of a rule-based labeler and then fine-tuned on a small set of expert annotations augmented with automated backtranslation. We find that our final model, CheXbert, is able to outperform the previous best rule-based labeler with statistical significance, setting a new SOTA for report labeling on one of the largest datasets of chest x-rays.
\end{abstract}

\section{Introduction}
The extraction of labels from radiology text reports enables important clinical applications, including large-scale training of medical imaging models \cite{wang_chestx-ray8_2017}. Many natural language processing systems have been designed to label reports using sophisticated feature engineering of medical domain knowledge \cite{pons_natural_2016}. On chest x-rays, the most common radiological exam, rule-based methods have been engineered to label some of the largest available datasets \cite{johnson_mimic-cxr-jpg_2019}. While these methods have generated considerable advances, they have been unable to capture the full diversity of complexity, ambiguity and subtlety of natural language in the context of radiology reporting.

\begin{figure}[t]
    \includegraphics[width=\columnwidth]{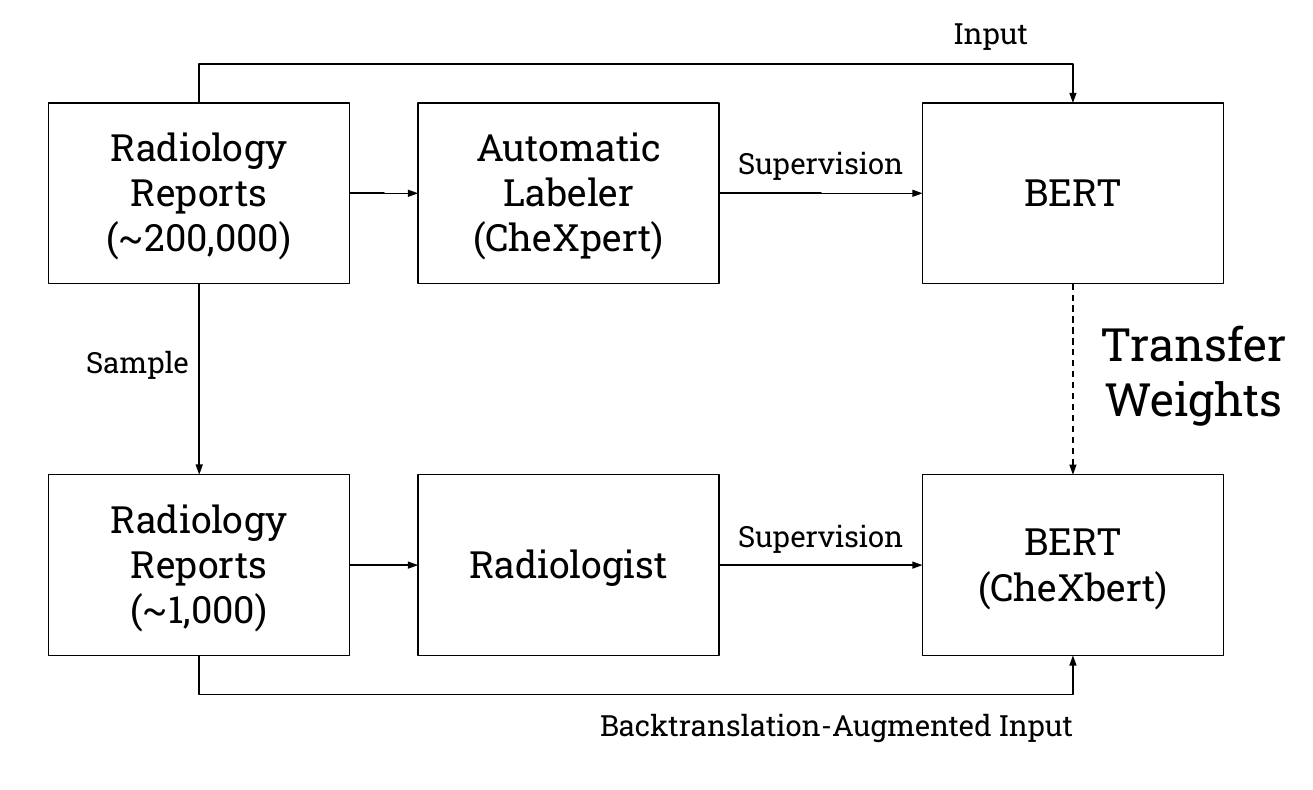}
    \caption{We introduce a method for radiology report labeling, in which a biomedically pretrained BERT model is first trained on annotations of a rule-based labeler, and then fine-tuned on a small set of expert annotations augmented with automated backtranslation.}
    \label{fig:main}
\end{figure}

More recently, Transformers have demonstrated success in end-to-end radiology report labeling \cite{drozdov_supervised_2020, wood_automated_2020}. However, these methods have shifted the burden from feature engineering to manual annotation, requiring considerable time and expertise for high quality. Moreover, these methods do not take advantage of existing feature-engineered labelers, which represent state-of-the-art on many medical tasks.

We introduce a simple method for gaining the benefits of both existing radiology report labelers and expert annotations to achieve highly accurate automated radiology report labeling. This approach begins with a biomedically pretrained BERT model \cite{devlin_bert_2019, peng_transfer_2019} trained on the outputs of an existing labeler, and performs further fine-tuning on a small corpus of expert annotations augmented with automated backtranslation. We apply this approach, shown in Figure \ref{fig:main}, to the task of radiology report labeling of chest x-rays, and call our resulting model \textit{CheXbert}.

CheXbert outperforms the previous best reported labeler \cite{irvin_chexpert_2019} on an external dataset, MIMIC-CXR \cite{johnson_mimic-cxr-jpg_2019}, with an improvement of 0.055 (95\% CI 0.039, 0.070) on the F1 metric, and is only 0.007 F1 away from a radiologist performance benchmark. We expect this method of training medical report labelers is broadly useful for natural language processing within the medical domain, where collection of expert labels is expensive, and feature engineered labelers already exist for many tasks.

\section{Related Work}
Many natural language processing systems have been developed to extract structured labels from free-text radiology reports \cite{pons_natural_2016, yadav_automated_2016, hassanpour_performance_2017, annarumma_automated_2019, savova_mayo_2010, wang_clinical_2018, chen_integrating_2018, bozkurt_automated_2019}. In many cases, these methods have relied on heavy feature engineering that include controlled vocabulary and grammatical rules to find and classify properties of radiological findings. NegEx \cite{chapman_simple_2001}, a popular component of rule-based methods, uses simple regular expressions for detecting negation of findings and is often used in combination with ontologies such as the Unified Medical Language System (UMLS) \cite{bodenreider_unified_2004}. NegBio \cite{peng_negbio_2017}, an extension to NegEx, utilizes universal dependencies for pattern definition and subgraph matching for graph traversal search, includes uncertainty detection in addition to negation detection for multiple pathologies in chest x-ray reports, and is used to generate labels for the ChestX-Ray14 dataset \cite{wang_chestx-ray8_2017}.

The CheXpert labeler \cite{irvin_chexpert_2019} improves upon NegBio on chest x-ray report classification through more controlled extraction of mentions and an improved NLP pipeline and rule set for uncertainty and negation extraction. The CheXpert labeler has been applied to generate labels for the CheXpert dataset and MIMIC-CXR \cite{johnson_mimic-cxr-jpg_2019}, which are amongst the largest chest x-ray datasets publicly available.

Deep learning approaches have also been trained using expert-annotated sets of radiology reports \cite{xue_fine-tuning_2019}. In these cases, training set size, often driving the performance of deep learning approaches, is limited by radiologist time and expertise. \citet{chen_deep_2017} trained CNNs with GloVe embeddings \cite{pennington_glove_2014} on 1000 radiologist-labeled reports for classification of pulmonary embolism in chest CT reports and improved upon the previous rule-based SOTA, peFinder \cite{chapman_document-level_2011}. \citet{bustos_padchest_2019} trained both recurrent and convolutional networks in combination with attention mechanisms on 27,593 physician-labeled radiology reports and apply their labeler to generate labels. More recently, Transformer-based models have also been applied to the task of radiology report labeling. \citet{drozdov_supervised_2020} trained classifiers using BERT \cite{devlin_bert_2019} and XLNet \cite{yang_xlnet_2020} on 3,856 radiologist labeled reports to detect normal and abnormal labels. \citet{wood_automated_2020} developed ALARM, an MRI head report classifier on head MRI data using BioBERT \cite{lee_biobert_2019} models trained on 1500 radiologist-labeled reports, and demonstrate improvement over simpler fixed embedding and word2vec-based \cite{mikolov_distributed_2013} models \cite{zech_natural_2018}. 

Our work is closely related to approaches to reduce the number of expert annotations required for training medical report labelers \cite{callahan_medical_2019, ratner_snorkel_2020, banerjee_radiology_2018}. A method of weak supervision known as data programming \cite{ratner_snorkel_2018} has seen successful application to medical report labeling: in this method, users write heuristic labelling functions that programmatically label training data. \citet{saab_doubly_2019} used data programming to incorporate labeling functions consisting of regular expressions that look for phrases in radiology reports, developed with the help of a clinical expert in a limited time window, to label for intracranial hemorrhage in head CTs. \citet{dunnmon_cross-modal_2019} demonstrated that in under 8 hours of cumulative clinician time, a data programming method can approach the efficacy of large hand-labeled training sets annotated over months or years for training medical imaging models, including chest x-ray classifiers on the task of normal / abnormal detection. Beyond data programming approaches, \citet{drozdov_supervised_2020} developed a fully unsupervised approach utilizing a Siamese Neural Network and Gaussian Mixture Models, reporting performance similar to the CheXpert labeler without requiring any radiologist-labeled reports on the simplified task of normal / abnormal detection. Concurrently developed to our work is the CheXpert++ labeler \cite{mcdermott2020chexpert}, which was trained on the outputs of the rule-based CheXpert labeler and showed improved performance after a single additional epoch of fine-tuning on expert-labeled report sentences. 

\begin{figure}[t!]
    \includegraphics[width=\columnwidth]{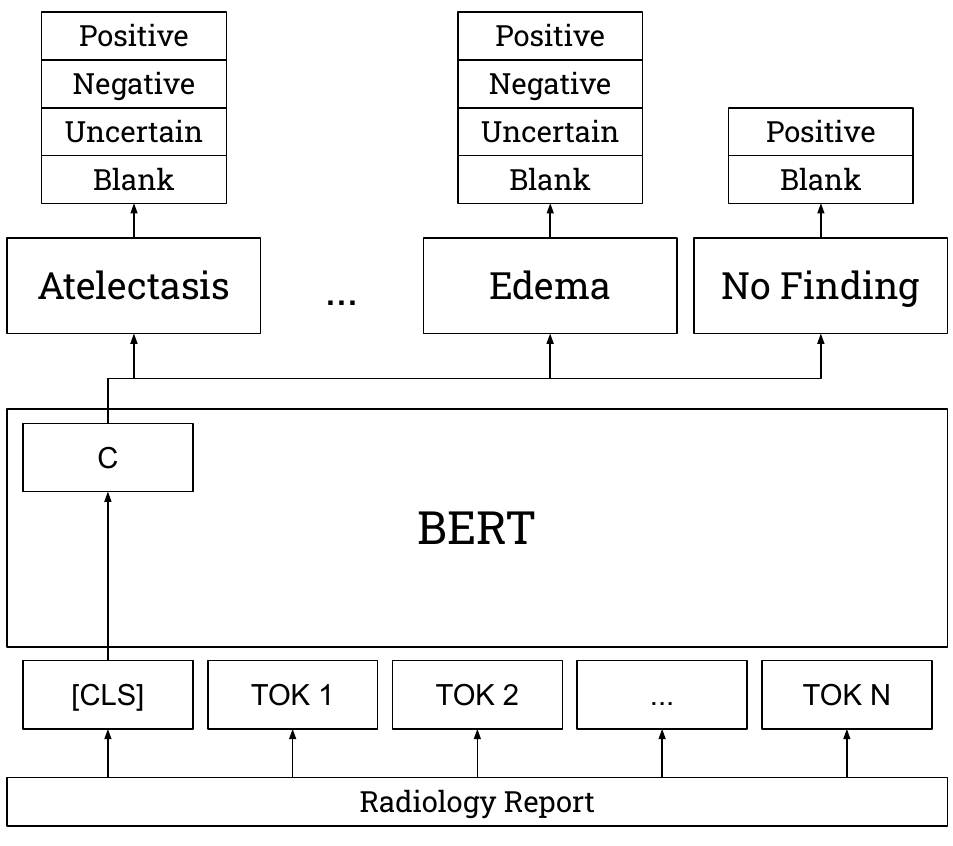}
    \caption{Model architecture. The model contains 14 linear heads, one for each medical observation, but only 3 heads are shown here.}
    \label{fig:architecture}
\end{figure}

\section{Methods}
\subsection{Task}
The report labeling task is to extract the presence of one or more clinically important observations (e.g. consolidation, edema) from a free-text radiology report.  More formally, a labeler takes in as inputs sentences from a radiology report and outputs for 13 observations one of the following classes: blank, positive, negative, and uncertain. For the 14th observation corresponding to \textit{No Finding}, the labeler only outputs one of the two following classes: blank or positive.

\subsection{Data}
Two existing large datasets of chest x-rays, CheXpert \cite{irvin_chexpert_2019} (consisting of 224,316 images), and MIMIC-CXR \cite{johnson_mimic-cxr-jpg_2019} (consisting of 377,110 images) are used in this study. Both datasets have corresponding radiology reports that have been labeled for the same set of 14 observations using the CheXpert labeler \cite{irvin_chexpert_2019}, from the \textit{Impression} section, or other parts of the radiology report.

A subset of both datasets also contain manual annotations by expert radiologists. On CheXpert, a total of 1000 reports (\textit{CheXpert manual set}) were reviewed by 2 board certified radiologists with disagreement resolution through consensus. On MIMIC-CXR, a total of 687 reports (\textit{MIMIC-CXR test set}) were reviewed by 2 board certified radiologists and manually labeled for the same 14 medical observations as in CheXpert. In this study, CheXpert is used for the development of models, and the MIMIC-CXR test set is used for evaluation.

Some reports from the same patient appear multiple times in the CheXpert dataset. Removing duplicate reports as well as the CheXpert manual set from the CheXpert dataset results in 190,460 reports, the class prevalences for which are shown in Table \ref{tab:chexpert_prevalences} of the Appendix. We remove excess spaces and newlines from all reports. 

\subsection{Model Architecture} All models use a modification of the BERT-base architecture \cite{devlin_bert_2019} with 14 linear heads (as shown in Figure \ref{fig:architecture}): 12 heads correspond to various medical abnormalities, 1 to medical support devices, and 1 to `No Finding'. Each radiology report text is tokenized, and the maximum number of tokens in each input sequence is capped at 512. The final-layer’s hidden state corresponding to the CLS token is then fed as input to each of the linear heads. 

\subsection{Training Details} For all our models, unless otherwise specified, we fine-tune all layers of the BERT model, including the embeddings, and feed the CLS token into the 14 linear heads to generate class scores for each medical observation.  BERT-Base contains $\sim110$ million parameters, and the linear heads contain $\sim40,000$ parameters.

All models are trained using cross-entropy loss and Adam optimization with a learning rate of $2 \times 10^{-5}$, as used in  \citet{devlin_bert_2019} for fine-tuning tasks. The cross-entropy losses for each of the 14 observations are added to produce the final loss. During training, we periodically evaluate our model on the dev set and save the checkpoint with the highest performance averaged over all 14 observations. All models are trained using 3 TITAN-XP GPUs with a batch size of 18.

\subsection{Evaluation} Models are evaluated on their average performance on three retrieval tasks: positive extraction, negative extraction, and uncertainty extraction. For each of the tasks, the class of interest (e.g. negative for the negative extraction and uncertain for the uncertainty extraction) is treated as the positive class, and the other classes are considered negative. For each of the 14 observations, we compute a weighted average of the F1 scores on each of the above three tasks, weighted by the support for each class of interest, which we call the \textit{weighted-F1} metric, henceforth simply abbreviated to F1.

We report the simple average of the F1 across all of the observations. We include the 95\% two-sided confidence intervals of the F1 using the nonparametric percentile bootstrap method with 1000 bootstrap replicates \cite{efron_bootstrap_1986}. 
\begin{table*}[t]
\centering
\fontfamily{cmss}\selectfont
\resizebox{\textwidth}{!}{%
\begin{tabular}{p{.3\linewidth} p{.35\linewidth} p{.35\linewidth}}
& Model                               & F1 (95\% CI) \\ \hline
\multirow{5}{*}{Training Strategy}                 & T-rad                      & 0.705 (0.680, 0.725) \\
                                                   & T.cls-rad                  & 0.286 (0.265, 0.305) \\
                                                   & T.token-rad                & 0.396 (0.374, 0.416) \\
                                                   & T-auto                     & 0.755 (0.731, 0.774) \\
                                                   & T-hybrid                   & 0.775 (0.753, 0.795) \\ \hline
\multirow{3}{*}{Biomedical Representations}        & Tbio-rad                   & 0.616 (0.587, 0.639) \\
                                                   & Tclinical-rad                & 0.677 (0.651, 0.699)  \\
                                                   & Tblue-rad                  & 0.741 (0.714, 0.763)
                             \\\hline
\multirow{4}{*}{\makecell[l]{With Backtranslation \\ Augmentation}} & T-rad-bt                   & 0.729 (0.702, 0.749)  \\
                                                   & T-hybrid-bt                & 0.795 (0.772, 0.815) \\
                                                   & Tblue-rad-bt               & 0.770 (0.747, 0.790) \\
                                                   & \textbf{Tblue-hybrid-bt (CheXbert)} & 0.798 (0.775, 0.816) \\ \hline
Previous SOTA                                      & CheXpert            & 0.743 (0.719, 0.764) \\ \hline
Benchmark                                          & Radiologist                & 0.805 (0.784, 0.823)           
\end{tabular}
}
\caption{\label{tab:headline} Average F1 score with 95\% confidence intervals for all our models, with comparisons to CheXpert labeler and radiologist benchmark.}
\end{table*}

\begin{table*}[t]
\centering
\fontfamily{cmss}\selectfont
\resizebox{\textwidth}{!}{%
\begin{tabular}{p{.3\linewidth}p{.35\linewidth}p{.35\linewidth}}
Category           & CheXbert             & Improvement over CheXpert            \\ \hline
Pneumonia         & 0.835 (0.789, 0.881) &  0.151 (0.093, 0.206)   \\
Fracture          & 0.791 (0.665, 0.895) &  0.120 (0.019, 0.236)   \\
Consolidation     & 0.877 (0.810, 0.935) &  0.105 (0.029, 0.192)  \\
Enlarged Cardiom. & 0.713 (0.623, 0.783) &  0.100 (0.038, 0.166)  \\
No Finding        & 0.640 (0.482, 0.759) &  0.097 (0.007, 0.182)  \\
Pleural Other     & 0.534 (0.372, 0.671)  &  0.056 (0.008, 0.124)   \\
Cardiomegaly      & 0.815 (0.759, 0.860) &  0.051 (0.018, 0.086)   \\
Pneumothorax      & 0.928 (0.892, 0.960) &  0.046 (0.015, 0.076)   \\
Atelectasis       & 0.940 (0.910, 0.971) &  0.023 (-0.001, 0.051)  \\
Support Devices   & 0.888 (0.856, 0.919) &  0.021 (0.004, 0.040)  \\
Edema             & 0.881 (0.843, 0.916) &  0.017 (-0.007, 0.042)  \\
Pleural Effusion  & 0.919 (0.892, 0.947) &  0.014 (-0.005, 0.034) \\
Lung Lesion       & 0.664 (0.550, 0.771) &  -0.019 (-0.098, 0.056)  \\
Lung Opacity      & 0.741 (0.684, 0.792) &  -0.021 (-0.056, 0.006) \\ \hline
Average           & 0.798 (0.775, 0.816) &  0.055 (0.039, 0.070)   \\ \hline
\end{tabular}%
}
\caption{\label{tab:chexbert_vs_chexpert} The F1 scores for CheXbert as well as improvements over the CheXpert labeler on the MIMIC-CXR test set, in descending order of improvement, and reported with 95\% confidence intervals.}
\end{table*}
\section{Experiments}

\subsection{Supervision Strategies}
We investigate models trained using three strategies: trained only on radiologist-labeled reports, trained only on labels generated automatically by the CheXpert labeler \cite{irvin_chexpert_2019}, and trained on a combination of the two.

\paragraph{Radiologist Labels} \textbf{T-rad} is obtained by training the model on the CheXpert manual set, fine-tuning all weights. As baselines, we also train models that freeze all weights in the BERT layers, and only update the weights in the linear heads: \textbf{T.cls-rad} is identical to T-rad in architecture, while \textbf{T.token-rad} averages the non-padding output tokens as the input into the linear heads rather than using the CLS token output. All models are trained using a random 75\%-25\% train-dev split on the CheXpert manual set, and are trained until convergence.
\paragraph{Automatic Labels} \textbf{T-auto} is obtained using labels generated by the rule-based CheXpert labeler, described in \citet{irvin_chexpert_2019}. T-auto is trained using a random 85\%-15\% train-dev split of the CheXpert dataset, different from the models trained on radiologist labels. T-auto is trained for 8 epochs, since slightly higher dev performance is observed compared to the typical 2-4 epochs for BERT fine-tuning tasks.
\paragraph{Hybrid Labels} \textbf{T-hybrid} is obtained by initializing a model with the weights of T-auto, and then fine-tuning it on radiologist-labeled reports, as for T-rad.

\paragraph{Results}
As shown in Table \ref{tab:headline},  T-rad achieves an F1 of $0.705$ $(0.680, 0.725)$, significantly higher than the performance of the baselines with T.cls-rad at $0.286$ $(0.265, 0.305)$, and T.token-rad at $0.396$ $(0.374, 0.416)$. T-auto achieves a higher F1 of $0.755$ $(0.731, 0.774)$. Superior performance is obtained by T-hybrid, with an F1 of $0.775$ $(0.753, 0.795)$.

\subsection{Biomedical Language Representations}
We investigate the effect of having models pre-trained on biomedical data. For the following models, we use an identical training procedure to T-rad, but initialize the weights differently. \textbf{Tbio-rad} is obtained by using BioBERT weight initializations \cite{lee_biobert_2019}. BioBERT was obtained by further pretraining the BERT weights on a large biomedical corpus comprising PubMed abstracts (4.5 billion words) and PMC full-text articles (13.5 billion words). \textbf{Tclinical-rad} is obtained by using Clinical BioBERT weight initializations \cite{alsentzer_publicly_2019}, which were obtained by further pretraining the BioBERT weights on 2 million clinical notes from the MIMIC-III database. Finally, \textbf{Tblue-rad} is obtained by using BlueBERT, a BERT model pretrained on PubMed abstracts and clinical notes (MIMIC-III) \cite{peng_transfer_2019}.

\paragraph{Results}
As shown in Table \ref{tab:headline}, Tbio-rad achieves an F1 of $0.616$ $(0.587, 0.639)$ and Tclinical-rad achieves an F1 of $0.677$ $(0.651, 0.699)$, lower than T-rad. However, Tblue-rad achieves an F1 of $0.741$ $(0.714, 0.763)$, higher than T-rad.
The drop in performance with Tbio-rad and Tclinical-rad may possibly be attributed to using different vocabulary, sequence length, and other configurations (stopping procedure, embedding dimensions) than those used by Tblue-rad, which uses the configurations provided in \citet{devlin_bert_2019}.

\subsection{Data Augmentation using Backtranslation}
We investigate the use of backtranslation to improve the performance of the models. Backtranslation is designed to generate alternate formulations of sentences by translating them to another language and back. Although backtranslation has been successfully used to augment text data in a variety of NLP tasks \citep{yu_qanet_2018, poncelas_investigating_2018}, to our knowledge, the technique is yet to be applied to a medical report extraction task. In this experiment, we augment the CheXpert manual set using Facebook-FAIR's winning submission to the WMT'19 news translation task \cite{ng_facebook_2019} to generate backtranslations. Although this submission includes models that produce German/English and Russian/English translations, initial experiments with Russian did not demonstrate semantically correct translations, so we only continued experiments with German. We use beam search with a beam size of 1 to select the single most likely translation. 
We perform this experiment using our best models: \textbf{Tblue-rad-bt} is obtained by using an identical training procedure to Tblue-rad on the augmented dataset (which is twice the size of the CheXpert manual set). \textbf{Tblue-hybrid-bt} is obtained by first training a BlueBERT-based labeler on automatically generated CheXpert labels, and then fine-tuning on radiologist-labeled reports of the CheXpert manual set, augmented by backtranslation. We also report the performance of T-rad-bt and T-hybrid-bt.

\paragraph{Results}
As shown in Table \ref{tab:headline}, T-rad-bt achieves an F1 score of $0.729$ $(0.702, 0.749)$, higher than that of T-rad. Similarly, T-hybrid-bt achieves an F1 of $0.795$ $(0.772, 0.815)$. Tblue-rad-bt achieves an F1 of $0.770$ $(0.747, 0.790)$, higher than that of the CheXpert labeler. Tblue-hybrid-bt achieves a superior F1 score of $0.798$ $(0.775, 0.816)$.

\subsection{Comparison to previous SOTA and radiologist benchmark}
We compare the performance of our best model to the previous best reported labeler, the CheXpert labeler \cite{irvin_chexpert_2019}, and to a radiologist benchmark. CheXpert is an automated rule-based labeler that extracts mentions of conditions like pneumonia by searching against a large manually curated list of words associated with the condition and then classifies mentions as uncertain, negative, or positive using rules on a universal dependency parse of the report. For the radiologist benchmark, the annotations by one of the 2 radiologists on the MIMIC-CXR test set is used, while the other is used as ground truth. We report the improvement of our best model, Tblue-hybrid-bt, which we also call \textbf{CheXbert}, over the CheXpert labeler by computing the paired differences in F1 scores on 1000 bootstrap replicates and provide the mean difference along with a 95\% two-sided confidence interval.


\paragraph{Results}
We observe that CheXbert has a statistically significant improvement $(p < 0.001)$ over the existing SOTA, CheXpert, which achieves a score of $0.743$ $(0.719, 0.764)$. Notably, we also find that Tblue-rad-bt, the best model trained only on manually labeled radiology reports, performs at least as well as the CheXpert labeler.

Table \ref{tab:chexbert_vs_chexpert} shows the F1 per class (along with 95\% confidence intervals) for CheXbert and for the improvements over CheXpert. CheXbert records an improvement in all but 2 medical conditions, and a statistically significant improvement in 9 of the 14 conditions. The largest improvements are observed for Pneumonia [$0.151$ $(0.093, 0.206)$], Fracture [$0.120$ $(0.019, 0.236)$], Consolidation [$0.105$ $(0.029, 0.192)$], Enlarged Cardiomediastinum [$0.100$ $(0.038, 0.166)$], and No Finding [$0.097$ $(0.007, 0.182)$]. Further significant improvements are observed for Pleural Other [$0.056$ $(0.008, 0.124)$], Cardiomegaly [$0.051$ $(0.018, 0.086)$], Pneumothorax [$0.046$ $(0.015, 0.076)$] and Support Devices [$0.021$ $(0.004, 0.040) $]. Overall, CheXbert achieves a statistically significant improvement on F1 of $0.055$ $(0.039, 0.070)$.
The board-certified radiologist achieves an F1 of $0.805$ $(0.784, 0.823)$, which is $0.007$ F1 points higher than the performance of CheXbert.

\paragraph{Training times} For all our models except the baselines, training on radiologist-labeled reports takes $\sim30$ minutes, training on the radiologist-labeled reports augmented via backtranslation takes $\sim50$ minutes. Training on the larger automatically labeled report set takes $\sim7$ hours.

\paragraph{Inference times} We benchmark the time taken by CheXbert and CheXpert to label all 190,460 report impressions in the CheXpert dataset. On a system with 32GB RAM and 1 CPU core, the CheXbert model takes $\sim 3.7$ hours. This is an order of magnitude faster than the $36$ hours required for CheXpert. With a single TITAN-XP GPU, the CheXbert model's inference time reduces to $\sim 18$ minutes.
\newcommand{\correct}[1]{#1 \ding{51}}%
\newcommand{\wrong}[1]{#1 \ding{55}}%

\begin{table*}[t]
\fontfamily{cmss}\selectfont
\resizebox{\textwidth}{!}{%
\begin{tabular}{|p{.5\linewidth}|p{.5\linewidth}|}
\hline
Report Segment and Labels & Reasoning \\ \hline

...two views of chest demonstrate \textit{cariomegaly} with no focal consolidation... 

\makecell[r]{
\\
Cardiomegaly\\ \wrong{CheXpert: Blank}\\ \correct{T-auto: Positive }
} & T-auto, in contrast to CheXpert, recognizes conditions with misspellings in the report like “cariomegaly” in place of “cardiomegaly”.

\\ \hline
...\textit{consistent with acute and/or chronic pulmonary edema}....  

\makecell[r]{
\\
Edema\\ \correct{CheXpert: Positive}\\ \wrong{T-auto: Uncertain}
}
& T-auto incorrectly detects uncertainty in the edema label, likely from the “and/or”; CheXpert correctly classifies this example as positive.

\\ \hline
...\textit{Normal heart size, mediastinal and hilar contours are unchanged in appearance}...

\makecell[r]{
\\
Enlarged Cardiomediastinum \\ \wrong{CheXpert: Negative}\\ \wrong{T-auto: Negative} \\ \correct{CheXbert: Uncertain}
}
& 
T-auto and CheXpert both incorrectly label this example as negative for enlarged cardiomediastinum; CheXbert correctly classifies it as uncertain, likely recognizing that ``unchanged'' is associated with uncertainty of the condition. The condition cannot be labeled positive or negative without more information.
\\ \hline
\end{tabular}
}
\caption{\label{tab:examples} Phrases from reports where CheXpert, T-auto, and CheXbert provide different labels. The correct label is indicated by a checkmark in the first column. The CheXpert versus T-auto comparisons are conducted on the CheXpert manual set. The CheXbert versus T-auto/CheXpert comparison is conducted on the MIMIC-CXR test set.}
\end{table*}

\section{Analysis}
\subsection{T-auto versus CheXpert}
We analyze whether T-auto, which is trained exclusively on labels from CheXpert (a rules-based labeler), can generalize beyond those rules.

We analyze specific examples in the CheXpert manual test set which T-auto correctly labels but CheXpert mislabels. On one example, T-auto is able to correctly detect uncertainty expressed in the phrase “cannot be entirely excluded,” which CheXpert is not able to detect because the phrase does not match any rule in its ruleset. Similarly, on another example containing “no evidence of pneumothorax or bony fracture,” T-auto correctly labels fracture as negative, while CheXpert labels fracture as positive since the phrasing does match any negation construct part of its ruleset. T-auto, in contrast to CheXpert, also recognizes conditions with misspellings in the report like “cariomegaly” in place of “cardiomegaly” and “mediastnium” in place of “mediastinum”. Examples of T-auto correctly labeling conditions mislabeled by CheXpert are provided in Table \ref{tab:t_auto_vs_chexpert} of the Appendix. Table \ref{tab:chexpert_vs_t_auto} of the Appendix contains examples of CheXpert correctly labeling conditions mislabeled by T-auto. An example of each case is shown in the top two rows of Table \ref{tab:examples}.

\subsection{CheXbert versus T-auto and CheXpert}

We analyze how CheXbert improves on T-auto and CheXpert using examples which CheXbert correctly labels but T-auto and CheXpert incorrectly label.

CheXbert is able to correctly detect conditions that CheXpert and T-auto are not able to. On one example, T-auto and CheXpert both mislabel a ``mildly enlarged heart'' as blank for cardiomegaly, while CheXbert correctly labels it positive. On another containing ``Right hilum appears slightly more prominent'' (an indicator for enlarged cardiomediastinum), CheXbert correctly classifies enlarged cardiomediastinum as positive, while T-auto and CheXpert do not detect the condition.

Furthermore, CheXbert correctly labels nuanced expressions of negation that both CheXpert and T-auto mislabel. On the example containing ``heart size is slightly larger but still within normal range,"  CheXpert and T-auto mistakenly label cardiomegaly as positive, while CheXbert correctly labels cardiomegaly as negative.
On another example containing the phrase ``interval removal of PICC lines”, CheXpert and T-auto detect ``PICC lines" as an indication of a support device but are unable to detect the negation indicated by ``removal", which CheXbert correctly does.

Additionally, CheXbert is able to correctly detect expressions of uncertainty that both CheXpert and T-auto mislabel. On an example containing “new bibasilar opacities, which given the clinical history are suspicious for aspiration,” CheXbert correctly identifies lung opacity as positive while CheXpert and T-auto incorrectly detect uncertainty (associating “suspicious” as a descriptor of “opacities”). More examples which CheXbert correctly labels but CheXpert and T-auto mislabel can be found in Table \ref{tab:chexbert_vs_rest} of the Appendix. A selected example is shown in the last row of Table \ref{tab:examples}.

\subsection{Report Changes with Backtranslation}
We analyze the phrasing and vocabulary changes that backtranslation introduces into the reports. Backtranslation frequently rephrases text. For instance, the sentence ``redemonstration of multiple right-sided rib fractures” is backtranslated to ``redemonstration of several rib fractures of the right side”. Backtranslation also introduces some error: the phrase “left costophrenic angle” is backtranslated to ``left costophrine angle” (``costophrine” is not a word), and the phrase ``left anterior chest wall pacer in place" is backtranslated to ``pacemaker on the left front of the chest wall", which omits the critical attribute of being in place. In many examples, the backtranslated text paraphrases medical vocabulary into possible semantic equivalents: “cutaneous" becomes ``skin", ``left clavicle" becomes ``left collarbone", ``osseous'' becomes ``bone" or ``bony", ``anterior" becomes ``front", and ``rib fracture" becomes ``broken ribs". More backtranslations with analyses are provided in Table \ref{tab:backtranslation} of the Appendix. Additionally, a physician validated that the backtranslation outputs used correct radiology language and maintained the semantics of the original report. The results are provided in Table \ref{tab:physician_backtrans} of the Appendix.

\section{Limitations}
Our study has several limitations. First, our hybrid/auto approaches require an already-existing labeler. Second, our report labeler has a maximum input token size of 512 tokens, but this may be easily extended to work with longer medical/radiology reports. In the CheXpert dataset, we found that only 3 of the 190,460 report impressions were longer than 512 tokens. Third, our task is limited to the 14 observations labeled for, and we do not test for the model's ability to label rarer conditions. However, CheXbert can mark No Finding as blank, which can indicate the presence of another condition if the other 13 conditions are also blank. Fourth, the ground truth labels for the MIMIC-CXR test set were determined by a single board-certified radiologist, and the use of more radiologists could demonstrate a truer comparison to the radiologist benchmark. Fifth, while we do test performance on a dataset from an institution unseen in training, additional datasets across institutions could be useful in further establishing the model's ability to generalize.
\section{Conclusion}
In this study, we propose a simple method for combining existing report labelers with hand-annotations for accurate radiology report labeling. In this method, a biomedically pretrained BERT model is first trained on the outputs of a labeler, and then further fine-tuned on the manual annotations, the set of which is augmented using backtranslation. We report five findings on our resulting model, CheXbert. First, we find that CheXbert outperforms models trained only on radiologist-labeled reports, or only on the existing labeler's outputs. Second, we find that CheXbert outperforms the BERT-based model not pretrained on biomedical data. Third, we find that CheXbert outperforms models which do not use backtranslation. Fourth, we find that CheXbert outperforms the previous best labeler, CheXpert (which was rules-based), with an improvement of 0.055 (95\% CI 0.039, 0.070) on the F1 metric; we also find that the best model trained only on manually labeled radiology reports (Tblue-rad-bt) performs at least as well as the CheXpert labeler. Fifth, we find that CheXbert is 0.007 F1 points from the radiologist performance benchmark, suggesting that the gap to ceiling performance is narrow.

We expect this method of training medical report labelers is broadly useful within the medical domain, where collection of expert labels can produce a small set of high quality labels, and existing feature engineered labelers can produce labels at scale. Extracting highly accurate labels from medical reports by taking advantage of both sources can enable many important downstream tasks, including the development of more accurate and robust medical imaging models required for clinical deployment.

\section*{Code Availability}

The code to train our CheXbert model and label reports, along with the trained CheXbert weights, is available in a public code repository: \href{https://github.com/stanfordmlgroup/CheXbert}{https://github.com/stanfordmlgroup/CheXbert}.

\section*{Acknowledgments} We would like to acknowledge the Stanford Machine Learning Group (\url{stanfordmlgroup.github.io}) and the Stanford Center for Artificial Intelligence in Medicine and Imaging (\url{AIMI.stanford.edu}) for infrastructure support. Thanks to Alistair Johnson for support in the radiologist benchmark, to Jeremy Irvin for support in the CheXpert labeler, Alex Tamkin for helpful comments, and Yifan Peng for helpful feedback.

\bibliography{emnlp2020}
\bibliographystyle{acl_natbib}

\onecolumn
\appendix
\setcounter{table}{0}
\renewcommand{\thetable}{A\arabic{table}}
\section{Physician validation of backtranslation quality}
\begin{table*}[h]
\centering
\caption{\label{tab:physician_backtrans} 
Physician validation of backtranslation output quality on a set of 100 randomly sampled reports from the CheXpert manual set and their backtranslations.}
\fontfamily{cmss}\selectfont
\begin{tabular}{|l|l|l|}
\hline
\textbf{Score} & \textbf{Valid radiology language} & \textbf{Preserves semantic information} \\ \hline
1     & 6                        & 14                   \\
2     & 48                       & 26                   \\
3     & 46                       & 60                   \\ \hline
\end{tabular}
\end{table*}

Although the CheXbert model shows empirical improvements using backtranslated reports, backtranslation can introduce additional noise into the reports. A physician validated the quality of the backtranslation outputs. For this experiment, we randomly sampled 100 reports from the CheXpert manual set. The physician read each original report and its backtranslation, and assigned a score for whether the backtranslation a) used valid radiology language, and b) maintained the semantics of the report. For each of tasks a) and b), the expert assigned a score of 1 (worst), 2 or 3 (highest).

For task a), a score of 3 means the backtranslation contained near-perfect radiology language, a 2 means the backtranslation had only minor deviations from valid radiology language, and 1 means the backtranslation had a major deviation from valid radiology language.

For task b), a score of 3 means the backtranslation fully preserved the semantics of the original, a 2 means the backtranslation contained minor semantic errors, and a 1 means the backtranslation had a major change or loss of semantic information compared to the original report.


\newpage
\setcounter{table}{0}
\renewcommand{\thetable}{B\arabic{table}}
\section{Additional results}
\begin{table*}[ht]
\fontfamily{cmss}\selectfont
\caption{\label{tab:chexpert_prevalences} After removing duplicate reports for the same patient from the CheXpert dataset (excluding the CheXpert manual set), we are left with a total of 190,460 reports. Labels for these reports are provided by the CheXpert labeler. The class prevalences of this set are displayed for each medical condition.}

\resizebox{\textwidth}{!}{%
\begin{tabular}{|l|l|l|l|l|}
\hline
\textbf{Condition}         & \textbf{Positive} & \textbf{Negative} & \textbf{Uncertain} & \textbf{Blank}    \\ \hline
Atelectasis                & 29,818 (15.66\%)  & 1,018 (0.53\%)    & 29,832 (15.66\%)   & 129,792 (68.15\%) \\ 
Cardiomegaly               & 23,302 (12.23\%)  & 7,809 (4.10\%)    & 6,682 (3.51\%)     & 152,667 (80.16\%) \\ 
Consolidation              & 12,977 (6.81\%)   & 19,397 (10.18\%)  & 24,345 (12.78\%)   & 133,741 (70.22\%) \\ 
Edema                      & 49,725 (26.11\%)  & 15,867 (8.33\%)   & 11,746 (6.17\%)    & 113,122 (59.39\%) \\ 
Enlarged Cardiomed. & 9,129 (4.79\%)    & 15,165 (7.96\%)   & 10,278 (5.40\%)    & 155,888 (81.85\%) \\ 
Fracture                   & 7,364 (3.87\%)    & 1,960 (1.03\%)    & 488 (0.26\%)       & 180,648 (94.85\%) \\ 
Lung Lesion                & 6,955 (3.65\%)    & 758 (0.40\%)      & 1,084 (0.57\%)     & 181,663 (95.38\%) \\ 
Lung Opacity               & 94,156 (49.44\%)  & 5,006 (2.63\%)    & 4,404 (2.31\%)     & 86,894 (45.62\%)  \\ 
No Finding                 & 16,795 (8.82\%)   & NA                & NA                 & 173,665 (91.18\%) \\ 
Pleural Effusion           & 77,028 (40.44\%)  & 25,097 (13.18\%)  & 9,565 (5.02\%)     & 78,770 (41.36\%)  \\ 
Pleural Other              & 2,481 (1.30\%)    & 210 (0.11\%)      & 1,801 (0.95\%)     & 185,968 (97.64\%) \\ 
Pneumonia                  & 4,647 (2.44\%)    & 1,851 (0.97\%)    & 15,907 (8.35\%)    & 168,055 (88.24\%) \\ 
Pneumothorax               & 17,688 (9.29\%)   & 47,566 (24.97\%)  & 2,704 (1.42\%)     & 122,502 (64.32\%) \\ 
Support Devices            & 107,601 (56.50\%) & 5,319 (2.79\%)    & 910 (0.48\%)       & 76,630 (40.23\%)  \\ \hline 
\end{tabular}%
}
\end{table*}
\begin{table*}[t]
\centering
\caption{\label{tab:dev_set} Dev set F1 scores for all our models. The dev set for all rad models and T-hybrid consists of 250 randomly sampled reports from the CheXpert manual set. The dev set for T-auto is a random 15\% split of the CheXpert dataset. The dev set for all models using backtranslation is obtained by augmenting the 250 randomly sampled reports from the CheXpert manual set by backtranslation. Tblue-hybrid-bt is first trained on labels generated by the CheXpert labeler, and then fine-tuned on radiologist labels augmented by backtranslation. Before fine-tuning on radologist labels, it obtains an F1 of 0.977 on the 15\% dev split of the CheXpert dataset.}
\fontfamily{cmss}\selectfont
\resizebox{\textwidth}{!}{%
\begin{tabular}{|p{.3\linewidth}| p{.35\linewidth}| p{.35\linewidth}|}\hline
& Model                               & F1 \\ \hline
\multirow{5}{*}{Training Strategy}                 & T-rad                      & 0.848 \\
                                                   & T.cls-rad                  & 0.411 \\
                                                   & T.token-rad                & 0.518 \\
                                                   & T-auto                     & 0.977 \\
                                                   & T-hybrid                   & 0.904 \\ \hline
\multirow{3}{*}{Biomedical Representations}        & Tbio-rad                   & 0.760  \\
                                                   & Tclinical-rad                & 0.802  \\
                                                   & Tblue-rad                  &  0.866
                             \\\hline
\multirow{4}{*}{\makecell[l]{With Backtranslation \\ Augmentation}} & T-rad-bt                   & 0.846  \\
                                                   & T-hybrid-bt                & 0.905 \\
                                                   & Tblue-rad-bt               & 0.865 \\
                                                   & Tblue-hybrid-bt (CheXbert) & 0.912 \\ \hline
\end{tabular}
}
\end{table*}
\begin{table*}[h]
\caption{\label{tab:counts_analysis} The differences in the number of times labels were correctly assigned by one model versus another model. For example, in the first column named “T-auto $>$ CheXpert,” we report the difference between the number of times T-auto correctly classifies a label and the number of times CheXpert correctly classifies a label. We record the differences between a pair of models by category (blank, positive, negative, uncertain) and by total. These occurrences are obtained on the MIMIC-CXR test set.}
\centering
\fontfamily{cmss}\selectfont
\resizebox{\textwidth}{!}{%

\begin{tabular}{|p{.1\linewidth}|p{.30\linewidth}|p{.30\linewidth}|p{.30\linewidth}|}
        \hline
          & \textbf{T-auto $>$ CheXpert} & \textbf{CheXbert $>$ CheXpert} & \textbf{CheXbert $>$ Radiologist} \\ \hline
Blank     & 0                            & 29                             & 56                            \\
Positive  & -22                          & 11                              & 56                            \\
Negative  & 14                           & 45                             & 9                            \\
Uncertain & 16                           & 46                             & -3                           \\ \hline
Total     & 8                            & 131                            & 118                           \\ \hline
\end{tabular}
}
\end{table*}
\begin{table*}[t]
\caption{\label{tab:t_auto_vs_chexpert} Examples where T-auto correctly assigns a label while CheXpert misassigns that label on the CheXpert manual set.  We include speculative reasoning for the classifications.}
\fontfamily{cmss}\selectfont
\resizebox{\textwidth}{!}{%
\begin{tabular}{|p{.70\linewidth}|p{.3\linewidth}|}
\hline
\textbf{Example \& Labels}                                                                   & \textbf{Reasoning}                                                                                                                                                                                          \\ \hline
...redemonstration of diffuse nodular air space opacities which are unchanged from prior examination \textbf{which may represent air space pulmonary edema versus infection}, as clinically correlated...

\makecell[r]{
\\
Edema \\ \wrong{CheXpert: Positive}\\ \correct{T-auto: Uncertain}
} & T-auto appears to detect uncertainties indicated by words like "may" and "versus" on conditions. In this case, this phrase did not match an uncertainty detection rule in the CheXpert classifier. \\ \hline
... there has been interval development of left basilar patchy airspace opacity, which likely represents atelectasis, \textbf{although consolidation cannot be entirely excluded}...

\makecell[r]{
\\
Consolidation \\ \wrong{CheXpert: Positive}\\ \correct{T-auto: Uncertain}
}
& Unlike CheXpert, T-auto correctly detects uncertainty conveyed in the phrase "cannot be entirely excluded".                                                                                        \\ \hline
1. no radiographic evidence of acute cardiopulmonary disease. 2. \textbf{no evidence of pneumothorax or bony fracture}.   

\makecell[r]{
\\
Fracture \\ \wrong{CheXpert: Positive}\\ \correct{T-auto: Negative}
}
& In this example, T-auto is able to detect a negation indicated by "no evidence of”. CheXpert is not able to pick up this negation construction as part of its ruleset.                             \\ \hline

\end{tabular}
}
\end{table*}
\begin{table*}[t]
\caption{\label{tab:chexpert_vs_t_auto}Examples where CheXpert correctly assigns a label while T-auto misassigns that label on the CheXpert manual set. We include speculative reasoning for the classifications.}
\fontfamily{cmss}\selectfont
\resizebox{\textwidth}{!}{%
\begin{tabular}{|p{.70\linewidth}|p{.3\linewidth}|}
\hline
\textbf{Example \& Labels}                                       & \textbf{Reasoning}                
\\ \hline
...2.\textbf{mild cardiomegaly}. persistent small bilateral pleural effusions, left greater than right...

\makecell[r]{
\\
Cardiomegaly \\ \correct{CheXpert: Positive}\\ \wrong{T-auto: Uncertain}
}
& T-auto mistakenly labels "mild cardiomegaly" as uncertain for cardiomegaly.                                                                                                       \\ \hline
...2.there are diffuse increased interstitial markings and prominence of the central vasculature, \textbf{consistent with acute and/or chronic pulmonary edema}...

\makecell[r]{
\\
Edema \\ \correct{CheXpert: Positive}\\ \wrong{T-auto: Uncertain}
}
& T-auto may have incorrectly detected uncertainty from “and/or,” which is a conjunction between “acute” and “chronic”.                                                             \\ \hline
\end{tabular}
}
\end{table*}
\begin{table*}[]
\caption{\label{tab:chexbert_vs_rest}Examples where CheXbert correctly assigns a label while both T-auto and CheXpert misassign that label on the MIMIC-CXR test set. We include speculative reasoning for the classifications.}
\fontfamily{cmss}\selectfont
\resizebox{\textwidth}{!}{%
\begin{tabular}{|p{.7\linewidth}|p{.3\linewidth}|}
\hline
\textbf{Example \& Labels}                                                                                                                                                    & \textbf{Reasoning}                                                                                                                                                                                                                                                                                                 \\ \hline
\textbf{New bibasilar opacities}, which given the clinical history are suspicious for aspiration, possibly developing pneumonia.    

\makecell[r]{
\\
Lung Opacity \\ \wrong{CheXpert: Uncertain}\\ \wrong{T-auto: Uncertain} \\ \correct{CheXbert: Positive}
}

& The word “suspicious” does not modify “opacities” in this sentence. Although CheXbert correctly identifies this, CheXpert and T-auto misclassify the “opacities” as uncertain.                                                 
\\ \hline
...Coalescent areas in the left upper and lower zones \textbf{could well reflect regions of consolidation}. The right lung is essentially clear...                 

\makecell[r]{
\\
Consolidation \\ \wrong{CheXpert: Positive}\\ \wrong{T-auto: Positive} \\ \correct{CheXbert: Uncertain}
}

& CheXbert correctly detects that consolidation is uncertain, as indicated by the phrase “could well reflect”.                                                               \\ \hline
Removal of dialysis catheter with no evidence of pneumothorax. \textbf{Heart is mildly enlarged} and is accompanied by vascular engorgement and new septal lines consistent with interstitial edema...                     

\makecell[r]{
\\
Cardiomegaly \\ \wrong{CheXpert: Blank}\\ \wrong{T-auto: Blank} \\ \correct{CheXbert: Positive}
}

& Due to a ruleset limitation, CheXpert only looks at “the heart” or “heart size” but not “heart” independently when checking for mentions of cardiomegaly. However, CheXbert recognizes mentions of cardiomegaly implied by phrases like “heart is mildly enlarged”.   \\ \hline 
No previous images. There is hyperexpansion of the lungs suggestive of chronic pulmonary disease. Prominence of engorged and ill-defined pulmonary vessels is consistent with the clinical diagnosis of pulmonary vascular congestion, though in the absence of previous images it is difficult to determine whether any this appearance could reflect underlying chronic pulmonary disease. \textbf{The possibility of supervening consolidation would be impossible to exclude} on this single study, especially without a lateral view. No evidence of pneumothorax.

\makecell[r]{
\\
Consolidation \\ \wrong{CheXpert: Positive}\\ \wrong{T-auto: Positive} \\ \correct{CheXbert: Uncertain}
}
& CheXbert correctly detects uncertainty for consolidation indicated by the word “possibility”. Both T-auto and CheXpert misclassify consolidation.                                                                                                                                                                                       \\ \hline
1. Left suprahilar opacity and fiducial seeds are again seen, although appears slightly less prominent/small in size, although as mentioned on the prior study, could be further evaluated by chest CT or PET-CT. 2. \textbf{Right hilum appears slightly more prominent} as compared to the prior study, which may be due to patient positioning, although increased right hilar lymphadenopathy is not excluded.      

\makecell[r]{
\\
Enlarged Cardiomediast. \\ \wrong{CheXpert: Blank}\\ \wrong{T-auto: Blank} \\ \correct{CheXbert: Positive}
}

& The right hilum appearing more prominent is an indicator of enlarged cardiomediastinum, which is clinically understood. If the hilum is growing, then the entire mediastinum is growing. Although both CheXpert and T-auto mislabeled this report impression, CheXbert successfully labeled it positive for enlarged cardiomediastinum. \\ \hline
\end{tabular}
}
\end{table*}
\begin{table*}[]
\fontfamily{cmss}\selectfont
\resizebox{\textwidth}{!}{%
\begin{tabular}{|p{.7\textwidth}|p{.3\textwidth}|}
\hline
\textbf{Example (cont.) \& Labels (cont.)}                                                                                                                                           & \textbf{Reasoning (cont.)}                                                                                                                                                                                                                            \\ \hline
As compared to the previous radiograph, there is no relevant change. The reduced volume of the right hemithorax with areas of lateral pleural thickening. The areas of pleural thickening are constant, size and morphology. Unchanged perihilar areas of fibrosis. \textbf{Unchanged size and aspect of the cardiac silhouette}, no pathologic changes in the left lung.

\makecell[r]{
\\
Cardiomegaly \\ \wrong{CheXpert: Positive}\\ \wrong{T-auto: Positive} \\ \correct{CheXbert: Uncertain}
}

& CheXbert correctly identifies uncertainty, as the cardiac silhouette is "unchanged," which means that it cannot be labeled positive or negative without additional information regarding the previous state. Both CheXpert and T-auto incorrectly label this example as positive for cardiomegaly.                        \\ \hline
AP chest compared to \_\_\_: Small-to-moderate left pleural effusion has increased slightly over the past several days. Moderate enlargement of the cardiac silhouette accompanied by mediastinal vascular engorgement is also slightly more pronounced. Pulmonary vasculature is engorged but there is no edema. Consolidation has been present without appreciable change in the left lower lobe since at least \_\_\_.  \textbf{Mediastinum widened} at the thoracic inlet by a combination of tortuous vessels and mediastinal fat deposition. Right jugular introducer ends just above the junction with left brachiocephalic vein.                    

\makecell[r]{
\\
Enlarged Cardiomediast. \\ \wrong{CheXpert: Blank}\\ \wrong{T-auto: Blank} \\ \correct{CheXbert: Positive}
}

& CheXbert correctly identifies enlarged cardiomediastinum from the phrase "mediastinum widened," which is a slightly different way of describing enlarged cardiomediastinum that CheXpert and T-auto both miss.                     \\ \hline
Moderately enlarged heart size, stable since \_\_\_. \textbf{No findings concerning for pulmonary edema} or pneumonia.                

\makecell[r]{
\\
Edema \\ \wrong{CheXpert: Uncertain}\\ \wrong{T-auto: Uncertain} \\ \correct{CheXbert: Negative}
}
&  Unlike T-auto and CheXpert, CheXbert correctly labels edema as negative, presumably understanding that the initial phrase ``no findings" applies to both edema and pneumonia. \\ \hline
AP chest compared to \_\_\_ and \_\_\_: As far as I can tell, given the severe anatomic distortion of the chest cage and its contents, lungs were clear on \_\_\_. Small region of opacification may have been developing lateral to the left hilus on \_\_\_, and today there is a suggestion of some new opacification at the base of the lung, but these observations are far from certain. I am not even confident that conventional radiographs, should the patient be able to cooperate for them, would clarify the issue. CT scanning, if feasible, would certainly confirm if the lungs are clear, but in the absence of a baseline study \textbf{it might be difficult to distinguish atelectasis from pneumonia}. Pleural effusion is minimal if any. Heart is probably not enlarged. Nasogastric tube is looped in the stomach. Right PIC line ends in the mid SVC. No pneumothorax.

\makecell[r]{
\\
Atelectasis \\ \wrong{CheXpert: Positive}\\ \wrong{T-auto: Positive} \\ \correct{CheXbert: Uncertain}
}
& The report states that ``it might be difficult to distinguish atelectasis from pneumonia" which indicates uncertainty, and this is correctly identified by CheXbert. CheXpert and T-auto simply label atelectasis as positive.                                                     \\ \hline                                                                                                                                                                                          
\end{tabular}
}
\end{table*}
\begin{table}[]
\fontfamily{cmss}\selectfont
\resizebox{\textwidth}{!}{%
\begin{tabular}{|p{0.7\linewidth}|p{.3\linewidth}|}
\hline
\textbf{Example (cont.) \& Labels (cont.)}                                                                                                                                     & \textbf{Reasoning (cont.)}                                                                                                                                                                                                                                                                                                                                                                                                                                           \\ \hline
Two frontal views of the chest show new mild interstitial pulmonary edema. Interval increase in mediastinal caliber therefore is probably due to distention of mediastinal veins. \textbf{Heart size is slightly larger but still within normal range}. Pleural effusions are minimal, if any. No focal pulmonary abnormality. No pneumothorax. ET tube is in standard placement and a nasogastric tube passes below the diaphragm and out of view.

\makecell[r]{
\\
Cardiomegaly \\ \wrong{CheXpert: Positive}\\ \wrong{T-auto: Positive} \\ \correct{CheXbert: Negative}
}
& Although CheXpert and T-auto mistakenly label cardiomegaly as positive given the phrase the "heart is slightly larger," the following phrase "but still within normal range" implies that cardiomegaly is negative. CheXbert correctly classifies this example as negative for cardiomegaly.          \\ \hline
As compared to the previous radiograph, the pre-existing right upper lobe \textbf{pneumonia is completely resolved}. The pre-existing signs of mild fluid overload, however, are still present. The pre-existing cardiomegaly is unchanged. Several calcified lung nodules are also unchanged. Unchanged alignment of the sternal wires. No acute pneumonia, no pleural effusions.              

\makecell[r]{
\\
Pneumonia \\ \wrong{CheXpert: Positive}\\ \wrong{T-auto: Positive} \\ \correct{CheXbert: Negative}
}
&  CheXbert correctly labels pneumonia as negative, as implied by the phrase " pneumonia is completely resolved," while CheXpert and T-auto both mislabel pneumonia as positive.                                                                                                                                                                                                                                                                                                                                                                                                                                                                          \\ \hline
Subsegmental right lung base atelectasis. Increasing loss of vertebral body height at T11. Stable L1 compression fracture. Right shoulder humeral DJD. \textbf{Interval removal of PICC lines}.                               

\makecell[r]{
\\
Support Devices \\ \wrong{CheXpert: Positive}\\ \wrong{T-auto: Positive} \\ \correct{CheXbert: Negative}
}
& CheXbert, presumably using a semantic understanding of the word ``removal", correctly labels support devices as negative. CheXpert and T-auto pick up on ``PICC lines" but do not detect the negation. Both incorrectly label support devices as positive.  

\\ \hline

AP chest compared to \_\_\_: Small-to-moderate left pleural effusion has increased slightly over the past several days. Moderate enlargement of the cardiac silhouette accompanied by mediastinal vascular engorgement is also slightly more pronounced. Pulmonary vasculature is engorged but there is no edema. Consolidation has been present without appreciable change in the left lower lobe since at least \_\_\_.  Mediastinum widened at the thoracic inlet by a combination of tortuous vessels and mediastinal fat deposition. \textbf{Right jugular introducer ends just above the junction with left brachiocephalic vein}.                        

\makecell[r]{
\\
Support Devices \\ \wrong{CheXpert: Blank}\\ \wrong{T-auto: Blank} \\ \correct{CheXbert: Positive}
}
& A jugular introducer is a support device that wasn't included in CheXpert's list of mentions for support devices. Consequently CheXpert and T-auto, which trains on CheXpert labels, both incorrectly label support devices as blank. CheXbert, however, correctly labels support devices as positive. 

\\ \hline
                                                                                                                                                                                                                                                                                           
\end{tabular}
}
\end{table}
\begin{table}[]
\fontfamily{cmss}\selectfont
\resizebox{\textwidth}{!}{%
\begin{tabular}{|p{0.7\linewidth}|p{.3\linewidth}|}
\hline
\textbf{Example (cont.) \& Labels (cont.)}                                                                                                                                     & \textbf{Reasoning (cont.)}                     
\\ \hline
1. Interval removal of the sternal wires with placement of new sternal closure devices, mediastinal staples and tubes. Lungs are well inflated with linear streaky opacities seen at the left base likely representing scarring and/or subsegmental atelectasis. No evidence of pulmonary edema, pneumothorax, pleural effusions or focal airspace consolidation to suggest pneumonia. \textbf{Slight lucency at the left apex is felt to be related to underlying emphysema rather than representing a pneumothorax}.                              

\makecell[r]{
\\
Pneumothorax \\ \wrong{CheXpert: Positive}\\ \wrong{T-auto: Positive} \\ \correct{CheXbert: Negative}
}
& CheXbert correctly labels pneumothorax as negative, as the radiologist notes that the observation is related to emphysema rather than pneumothorax. In this complex negation, T-auto and CheXpert incorrectly label pneumothorax as positive.

\\ \hline                                                                                                                                                                                                                                                                                                                                                                                                                        
\end{tabular}
}
\end{table}
\begin{table*}[]
\caption{\label{tab:backtranslation} Examples of additional data samples generated using backtranslation on radiologist-annotated reports from the CheXpert manual set. Augmenting our relatively small set of radiologist-annotated reports with backtranslation proved useful in improving performance of our labeler on the MIMIC-CXR test set.}
\fontfamily{cmss}\selectfont
\resizebox{\textwidth}{!}{%
\begin{tabular}{|p{0.33\textwidth}|p{0.33\textwidth}|p{0.33\textwidth}|}
\hline
\textbf{Original Report} &
  \textbf{Backtranslation} &
  \textbf{Changes} \\ \hline
1. marked cardiomegaly with a configuration that raises concern for a pericardial effusion. possible mild edema. \newline \newline 2. healed left-sided rib fractures. &
  1. pronounced cardiomegaly with a configuration that raises concerns about a pericardial effusion. possible mild edema. \newline \newline 2. healed left-sided rib fractures. &
  “marked” is changed to the synonym “pronounced”, and “raises concern for” is rephrased as “raises concerns about”. \\ \hline
1. redemonstration of right side pleural effusion and bibasilar atelectasis unchanged from comparison. \newline \newline 2. redemonstration of multiple right-sided rib fractures. &
  1. redemonstration of the pleural effusion of the right side and the bibasilar atelectasia unchanged compared to the comparison. \newline \newline 2. redemonstration of several rib fractures of the right side. &
  “right side pleural effusion” is rephrased as “pleural effusion of the right side”, “unchanged from comparison” is rephrased to “compared to the comparison” and “multiple right-sided rib fractures” is rephrased as “several rib fractures of the right side”. \newline \newline However, “atelectasis” is incorrectly changed to “atelectasia”. \\ \hline
1. single ap portable semiupright view of the chest demonstrates no change in medical support devices. \newline \newline 2. persistent dense retrocardiac opacity and small to moderate left pleural effusion present. right perihilar opacity appears resolved. \newline \newline 3. stable cardiomediastinal silhouette. no pulmonary edema. \newline \newline 4. multilevel degenerative changes of the spine. &
  1. single ap portable semi-upright view of the breast showing no change in medical aids. \newline \newline 2. persistent dense retrocardiac opacity and small to moderate left pleural effusion presented. right perihilar opacity appears resolved. \newline \newline 3. stable cardiomediastinal silhouette. no pulmonary edema. \newline \newline 4. multi-level degenerative changes of the spine. &
  “semiupright” becomes “semi-upright”, “medical support devices” is changed to “medical aids”, “present” is changed to “presented” and “multilevel” is changed to “multi-level”. \newline \newline However, “chest” is incorrectly changed to “breast”. \\ \hline

\end{tabular}
}
\end{table*}
\begin{table}[]
\fontfamily{cmss}\selectfont
\resizebox{\textwidth}{!}{%
\begin{tabular}{|p{0.33\textwidth}|p{0.33\textwidth}|p{0.33\textwidth}|}
\hline
\textbf{Original Report (cont.)} &
  \textbf{Backtranslation (cont.)} &
  \textbf{Changes (cont.)} \\ \hline
1. single frontal view of the chest demonstrates a surgical drain projecting over the neck, a tracheostomy tube, a feeding tube which extends below the diaphragm and beyond the inferior margin of the film. cutaneous staples project over the left clavicle, and surgical clips are seen within the left neck. no evidence of pneumothorax. \newline \newline 2. a dense retrocardiac opacity may represent atelectasis versus consolidation.,small bilateral pleural effusions are present. A convex opacity at the right paratracheal region is of uncertain significance; recommend upright pa and lateral for further evaluation when the patient is able. \newline \newline 3. the cardiomediastinal silhouette and pulmonary vasculature are unremarkable. &
  1. a single frontal view of the breast shows a surgical drain extending over the neck, a tracheostolic tube, a feeding tube extending under the diaphragm and over the lower edge of the film. skin clamps protrude over the left collarbone, and surgical clips are visible in the left cervical area. no indication of pneumothorax. \newline \newline 2. dense retrocardiac opacity may represent ateltasia versus consolidation. small bilateral pleural effusions are present. convex opacity in the right paratracheal area is of uncertain importance; recommend upright pa and lateral for further assessment if the patient is able to do so. \newline \newline 3. the cardiastinal silhouette and pulmonary vasculature are unobtrusive. &
  “demonstrates a surgical drain projecting over” rephrased to “shows a surgical drain extending over”, “a feeding tube which extends below the diaphragm and beyond the inferior margin of the film” rephrased to “a feeding tube extending under the diaphragm and over the lower edge of the film”, “surgical clips are seen within the left neck” changed to the semantically equivalent “surgical clips are visible in the left cervical area”, “region is of uncertain significance” rephrased as “area is of uncertain importance”, “further evaluation when the patient is able” is rephrased as “further assessment if the patient is able to do so”, and “pulmonary vasculature are unremarkable” is changed to the semantically close “pulmonary vasculature are unobtrusive”. \newline \newline However “chest” incorrectly changed to “breast, “tracheostomy tube” incorrectly changed to “tracheostolic tube”, “cutaneous staples project over the left clavicle” changed to the semantically similar “skin clamps protrude over the left collarbone”, but “skin clamps” is suboptimal, “atelectasis” incorrectly changed to “ateltasia”, “cardiomediastinal” is incorrectly changed to “cardiastinal”. \\ \hline
\end{tabular}
}
\end{table}
\begin{table}[]
\fontfamily{cmss}\selectfont
\resizebox{\textwidth}{!}{%
\begin{tabular}{|p{0.33\textwidth}|p{0.33\textwidth}|p{0.33\textwidth}|}
\hline
\textbf{Original Report (cont.)} &
  \textbf{Backtranslation (cont.)} &
  \textbf{Changes (cont.)} \\ \hline
1. single ap view of the chest demonstrates hyperinflation of the lungs. \newline \newline 2. there are prominent interstitial opacities which are stable. there is a residual tiny left apical pneumothorax without interval change. \newline \newline 3. cardiomediastinal silhouette is stable. \newline \newline 4. there is nonvisualization of the left costophrenic angle limiting its evaluation and if concerned, repeat study can be performed. &
  1. a single view of the breast shows hyperinflation of the lungs. \newline \newline 2. there are prominent interstitial opacities that are stable. there is a remaining tiny left apical pneumothorax without interval change. \newline \newline 3. the cardiomediastinal silhouette is stable \newline \newline 4. there is no visualization of the left costophrine angle that restricts its assessment, and if affected, a repeat study can be conducted. &
  “demonstrates hyperinflation” is rephrased as “shows hyperinflation”, “residual” is changed to the synonym “remaining”, and “angle limiting its evaluation and if concerned, repeat study can be performed” is rephrased to “angle that restricts its assessment, and if affected, a repeat study can be conducted”. The replacement of “concerned” with “affected” appears suboptimal. \newline \newline However, “ap” is incorrectly removed from the phrase “single ap view of the chest”, “chest” is incorrectly changed to “breast”, and ”costophrenic angle” is incorrectly changed to “cortophrine angle”. \\ \hline
\end{tabular}}
\end{table}

\end{document}